\documentclass[11pt]{article}

\usepackage[final]{acl}

\usepackage{booktabs}
\usepackage{multirow}
\usepackage{tabularx}

\usepackage{times}
\usepackage{latexsym}
\usepackage{amsmath}

\usepackage[T1]{fontenc}

\usepackage[utf8]{inputenc}

\usepackage{microtype}

\usepackage{inconsolata}

\usepackage{graphicx}

\title{Towards Robust Calibration Dataset for Domain-General Mixture of Experts Pruning}

\author{
 \textbf{Luis Frentzen Salim\textsuperscript{1,2}},
 \textbf{Chia-Chun Wu\textsuperscript{1}},
 \textbf{Tran Van Nhiem\textsuperscript{3}},
 \textbf{Lun-Wei Ku\textsuperscript{1}},
 \textbf{Yung-Hui Li\textsuperscript{3}},
\\
 \textsuperscript{1}Institute of Information Science, Academia Sinica,
\\
 \textsuperscript{2}National Taiwan University of Science and Technology,
\\
 \textsuperscript{3}AI Research Center, Hon Hai Research Institute,
\\
 \small{
   \textbf{Correspondence:} \href{mailto:luisfrentzen@gmail.com}{luisfrentzen@gmail.com}
 }
}

\begin{document}
\maketitle
\begin{abstract}
Post-training expert pruning is a promising deployment technique for Mixture-of-Experts (MoE) models. The calibration set is an integral part of the pruning process, however, its correlation with evaluation performance is largely misunderstood and unintuitive. In this work, we present an empirical study of the robustness of calibration perturbations, specifically in MoE models. Showing how calibration size, language, and data distribution affect layer-wise expert pruning for gpt-oss-20b. Our findings reveal that (1) domain-general pruning quality receives minimal benefit from scaling calibration data beyond small sample sizes, (2) cross-lingual calibration transfers effectively, with English-only calibration maintaining strong performance across seven languages of varying typologies and scripts, and (3) a diverse calibration data resembling the pretraining distribution can yield comparable or superior performance to test-distribution data. Ultimately, we provide practical insights as a recipe for better understanding calibration for more effective domain-general expert pruning.
\end{abstract}

\section{Introduction}

Mixture of Experts (MoE) models are rising in popularity due to their sparse and efficient inference \cite{jiang2024mixtralexperts, dai2024deepseekmoeultimateexpertspecialization}. MoE models achieve state-of-the-art results while preserving computational efficiency by activating only a set of "experts" for every incoming token. Nonetheless, the substantial memory requirements and operational complexities of serving sparse models make production deployment expensive. Post-training expert pruning became the go-to strategy to compress MoE models and remove redundant experts \cite{chen2022taskspecificexpertpruningsparse,li2025slimmoestructuredcompressionlarge,xie2024moeprunerpruningmixtureofexpertslarge}. A central yet often misunderstood component of this process is the calibration set, a specific group of sequences used to assess expert relevance and decide which parameters to retain.

In this work, we investigate the robustness of the standard expert-pruning strategy across different calibration settings to provide practical insights for model expert pruning. Especially how changes in language, distribution, or set size will interact with the pruning result and downstream task performance. We conduct an empirical study using OpenAI's \emph{GPT-OSS} MoE models \cite{openai2025gptoss120bgptoss20bmodel}. More precisely, we ask the questions: (1) Considering its sparse nature, what is the extent of pruning experts in the MoE architecture? (2) How does calibration set size affect downstream task performance? (3) Can cross-lingual calibration help mitigate certain data limitations during expert pruning? (4) What is the optimal calibration distribution?

This work is motivated by a real-world deployment setting in which final evaluation distribution is not known, and calibration data is limited by size, domain coverage, and multilingual availability. In such settings, teams often cannot assume access to large in-domain or language-matched calibration sets. Ultimately, we provide practical recommendations regarding tuning calibration to pruning MoE experts for domain-general use cases. We present various calibration settings and their effects on pruning results, showing how to minimize the effort required to design a calibration set that ensures optimal deployment.

\section{Related Works}

\subsection{Mixture of Experts}

\paragraph{Sparse MoE architectures}
Mixture-of-Experts models scale model capacity by activating only a small subset of parameters per token \cite{fedus2022switchtransformersscalingtrillion, jiang2024mixtralexperts,dai2024deepseekmoeultimateexpertspecialization}.
In typical sparse MoE layers, a learned router assigns each token to the top-$k$ experts, yielding high parameter counts while keeping per-token compute closer to a dense model of the same active size \cite{fedus2022switchtransformersscalingtrillion, jiang2024mixtralexperts}.
This design has enabled training and serving of very large language models by improving compute--quality trade-offs and by allowing specialization across experts.

\paragraph{Routing behavior and specialization}
Prior work has observed that expert utilization is often highly imbalanced, with a small number of experts receiving a large fraction of tokens \cite{shazeer2017outrageouslylargeneuralnetworks, chi2022representationcollapsesparsemixture, jiang2024mixtralexperts}. Such an imbalance motivates analyses of expert redundancy and specialization, as well as methods to stabilize routing and encourage more uniform load distribution. These observations are directly relevant to our setup, so if certain experts are rarely selected or contribute only marginally to the final representation, they may be candidates for removal without significant degradation in quality.

\subsection{Model Pruning}

\paragraph{Pruning for efficiency}
Model pruning aims to reduce inference cost by removing parameters, structures, or computation while preserving task performance \cite{frankle2019lotterytickethypothesisfinding}.
In language models, common strategies include unstructured weight pruning, structured pruning (e.g., heads, neurons, channels), and layer dropping \cite{see-etal-2016-compression, huang2024largelanguagemodelpruning}.
Structured approaches are typically favored for deployment because they lead to predictable speedups on standard hardware.

Many pruning methods rely on a calibration set to estimate parameter or component importance via activation statistics, gradients, or sensitivity metrics. The choice of calibration data can meaningfully affect which components are deemed redundant, which in turn influences the final quality--efficiency trade-off.

\subsection{Expert Pruning}

\paragraph{Pruning in MoE models}
Expert pruning targets sparse MoE layers by removing entire experts, reducing memory footprint, and potentially improving latency by shrinking routing and expert computation \cite{lu-etal-2024-experts}.
Unlike dense-model pruning, expert pruning must account for the router: disabling an expert changes the routing distribution and can shift load to remaining experts \cite{xie2024moeprunerpruningmixtureofexpertslarge, li2025slimmoestructuredcompressionlarge}.
As a result, effective approaches typically combine an importance metric derived from routing or expert activations and a masking strategy that re-normalizes routing over surviving experts.

\paragraph{Robustness considerations}
Because MoE experts may specialize, pruning decisions can be sensitive to the calibration distribution.
Evaluating pruning under cross-domain and cross-lingual calibration helps characterize whether the importance estimates capture general utility or dataset-specific behavior, and whether pruning disproportionately harms certain capabilities.

\section{Methodology}
\subsection{Model}

We study expert pruning in sparse MoE language models, where each token is routed to a small subset of experts. Our base model is OpenAI's 20 billion parameters open-weight \emph{GPT-OSS} MoE. This model family specifically enables us to develop and validate the pruning pipeline on the smaller model for rapid iteration, then transfer the same procedure to the larger variant with minimal changes. 

\subsection{Expert Pruning}

\paragraph{Expert importance and reconstruction objective}
We follow the post-training expert pruning procedure of \citet{lu-etal-2024-experts}. At a high level, the method selects, \emph{for each MoE layer independently}, a subset of $r$ experts (out of $n$) that best preserves the original layer's behavior on a small calibration set.

Concretely, given a calibration set $\mathcal{C}$, we run the unpruned model and cache the input--output activations for each MoE layer. For a candidate subset $C$ with $|C|=r$, we form a pruned version of the layer by masking experts not in $C$ and their associated router weights, re-normalizing the router distribution over the remaining experts, and recomputing the layer output on the cached inputs. We score each subset by a reconstruction loss between the original output and the pruned-layer output (e.g., Frobenius norm / $\ell_2$ distance over token representations), and select the subset that minimizes this loss.
The final pruned checkpoint is obtained by concatenating the selected subsets across layers, without any parameter updates.

\paragraph{Pruning policy}
We prune in a layer-wise manner and preserve a fixed number of experts $r$ per MoE layer.\footnote{In practice, exhaustive enumeration of all $\binom{n}{r}$ subsets is intractable for large $n$; we use the same heuristic subset search as \citet{lu-etal-2024-experts}.}
Pruned experts are permanently removed (or equivalently masked) from the routing distribution at inference time, reducing the model's static parameters and memory footprint.

\paragraph{Calibration data}
We vary the calibration distribution and size to study how sensitive the importance estimates are to the chosen data.
In addition to in-distribution calibration, we consider cross-domain and cross-lingual calibration to probe robustness. To reflect our deployment setting, we additionally evaluate calibration on an anonymized internal Traditional Chinese corpus. We use this corpus solely as a calibration source and compare it with public proxy corpora to assess whether private in-domain data is necessary for effective pruning.

\subsection{Experimental Design}
\label{sec:exp_design}
In all experiments, we prune with the specified calibration configuration and then evaluate the resulting checkpoint on the same fixed evaluation suite. The default calibration set size is $64$, and the batch size is $16$. All experiments will follow these default values unless stated otherwise.

\paragraph{(1) How many experts should we keep?}
To study the accuracy--compression trade-off and identify a deployment-optimal region, we sweep over the retained-expert fraction (equivalently, the per-layer preserved-expert count $r$) and evaluate each pruned checkpoint.

\paragraph{(2) How much calibration data is enough?}
To test data efficiency, we fix a representative pruning ratio, vary the set size from very small to larger subsets, and report downstream task performance.

\paragraph{(3) Does calibration language need to match the deployment language?}
To evaluate cross-lingual robustness, we calibrate pruning using data in one language and evaluate on benchmarks in other languages. Cross-lingual calibration is an important aspect of our exploration, as the ability to efficiently transfer settings between languages would benefit certain languages, such as Traditional Chinese. In our setting, we use English as the calibration language and evaluate on 7 languages, varying in script and typology: English, Arabic, German, French, Japanese, Russian, and Chinese.

\paragraph{(4) Should calibration be pretraining-like or task-specific?}
To characterize the calibration trade-off most relevant to practice, we compare pruning calibrated on (i) pretraining-like, broad text and (ii) domain-specific data.
We prune under a fixed configuration and compare downstream performance to understand when general, pretraining-like calibration is sufficient and when task-specific calibration provides meaningful gains.

\subsection{Dataset}

We use a mixture of publicly available corpora and a private dataset for pruning calibration. 
The calibration sources are chosen to represent three practical regimes: broad pretraining-like data, task-specific data, and language-matched evaluation data. 
This design allows us to test whether expert-pruning decisions are primarily driven by general distributional coverage, downstream task similarity, or calibration-language matching. 
All calibration samples are used only to estimate expert importance during pruning; the pruned checkpoints are then evaluated on the fixed benchmark suite described in Section~\ref{sec:exp_design}. Our calibration sources are:

\paragraph{C4} \cite{2019t5} We use C4 as a public, broad-coverage English web corpus that approximates a pretraining-like calibration distribution. C4 serves as a great domain-general baseline as it often used for model pretraining and exposes the model to broad and diverse topics.

\paragraph{MATH} \cite{hendrycksmath2021} We use the MATH dataset as the mathematical calibration pool. The dataset contains competition-level mathematical problems and their solutions, making it a great fit for our calibration tests of domain specificity. 

\paragraph{MMLU-ProX Lite} \cite{xuan2025mmluproxmultilingualbenchmarkadvanced} For the cross-lingual calibration experiments in Section~\ref{sec:results}, we use MMLU-ProX validation examples as language-specific calibration data. This dataset lets us test whether matching the calibration language to the evaluation language is necessary for preserving multilingual performance.

\paragraph{Private Corpus} We also sample from our proprietary Traditional Chinese dataset, which better reflects our production language. The corpus is a highly curated, multiple-domain dataset originally optimized for continued pretraining and alignment. Prior to processing, the raw aggregated data amounted to roughly 141.13B tokens (421.24 GB). 

Since manual curation at this scale is infeasible, we use highly capable LLMs as agents to clean, filter low-quality documents, and perform domain classification. Reducing the dataset to a highly refined 97.71 billion tokens. This final corpus is classified into a 26-domain taxonomy (e.g., Arts, Finance, Law \& Government, Computers \& Electronics). Furthermore, to increase the diversity and richness of the data, we follow the CO-STAR data augmentation framework. Raw domain data is passed through multiple proprietary LLMs using varied configurations to generate diverse rewrites, expansions, and structural variations and finally LLM-judged to ensure data quality. The full dataset will be made available upon publication to support community research and development.

\begin{figure}[t]
  \includegraphics[width=\columnwidth]{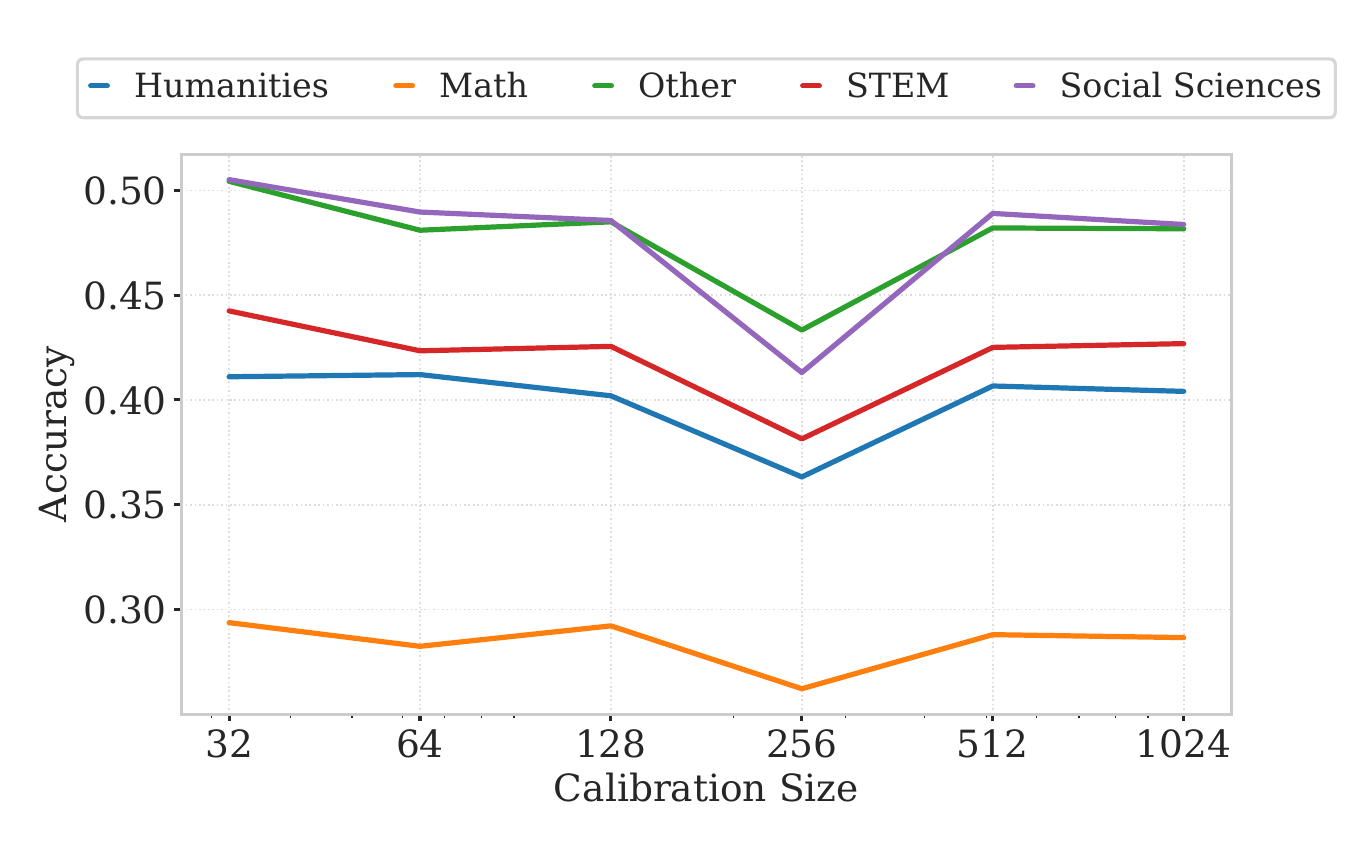}
  \caption{Evaluation performance across different calibration set sizes shows that scaling up calibration doesn't necessarily yield better post-prune performance. We fixed the number of retained experts to be $24$.}
  \label{fig:calib}
\end{figure}

\subsection{Evaluation}

To evaluate expert pruning performance, we use the LM Evaluation Harness as our primary evaluation framework. We evaluate on the following tasks:

\begin{itemize}
    \item \textbf{MathQA} \cite{amini-etal-2019-mathqa} A math benchmark that includes natural-language questions paired with quantitative reasoning and arithmetic.
    We include it as an additional math-heavy signal complementary to GSM8K to probe whether pruning affects performance differently across reasoning tasks.
    \item \textbf{GSM8K} \cite{cobbe2021gsm8k} Grade-school math word problems that test multi-step arithmetic reasoning and instruction following.
    We report accuracy under strict final-answer matching and analyze how pruning affects chain-of-thought-style reasoning.
    \item \textbf{MMLU} \cite{hendrycks2021mmlu} A broad multiple-choice benchmark spanning dozens of academic subjects, designed to probe factual knowledge and problem solving.
    We use it as a general capability check to ensure that pruning does not disproportionately degrade performance on knowledge-heavy tasks. For analysis, we aggregate MMLU subject-level results into coarse categories to reduce variance and highlight domain-level trends.
    \item \textbf{MMLU-ProX Lite} \cite{xuan2025mmluproxmultilingualbenchmarkadvanced} A harder, more reasoning-oriented multiple-choice suite derived from the MMLU family.
    We include it to stress-test pruning on more challenging questions, where small representation shifts can cause larger drops in accuracy. For cross-lingual analysis and visualization, we group each language into its own category.
\end{itemize}

\subsection{Experimental Setup}

All the experiments were conducted on H100 GPUs. We leverage the standard MXFP4 quantization from \emph{GPT-OSS} to run each model efficiently. We run the models with the vLLM library \cite{kwon2023efficient}, and all evaluations are conducted with the LM Evaluation Harness \cite{eval-harness}. We run each evaluation in three iterations with different random seeds and report the average.

\section{Results and Discussion}
\label{sec:results}

\begin{figure}[t]
  \includegraphics[width=\columnwidth]{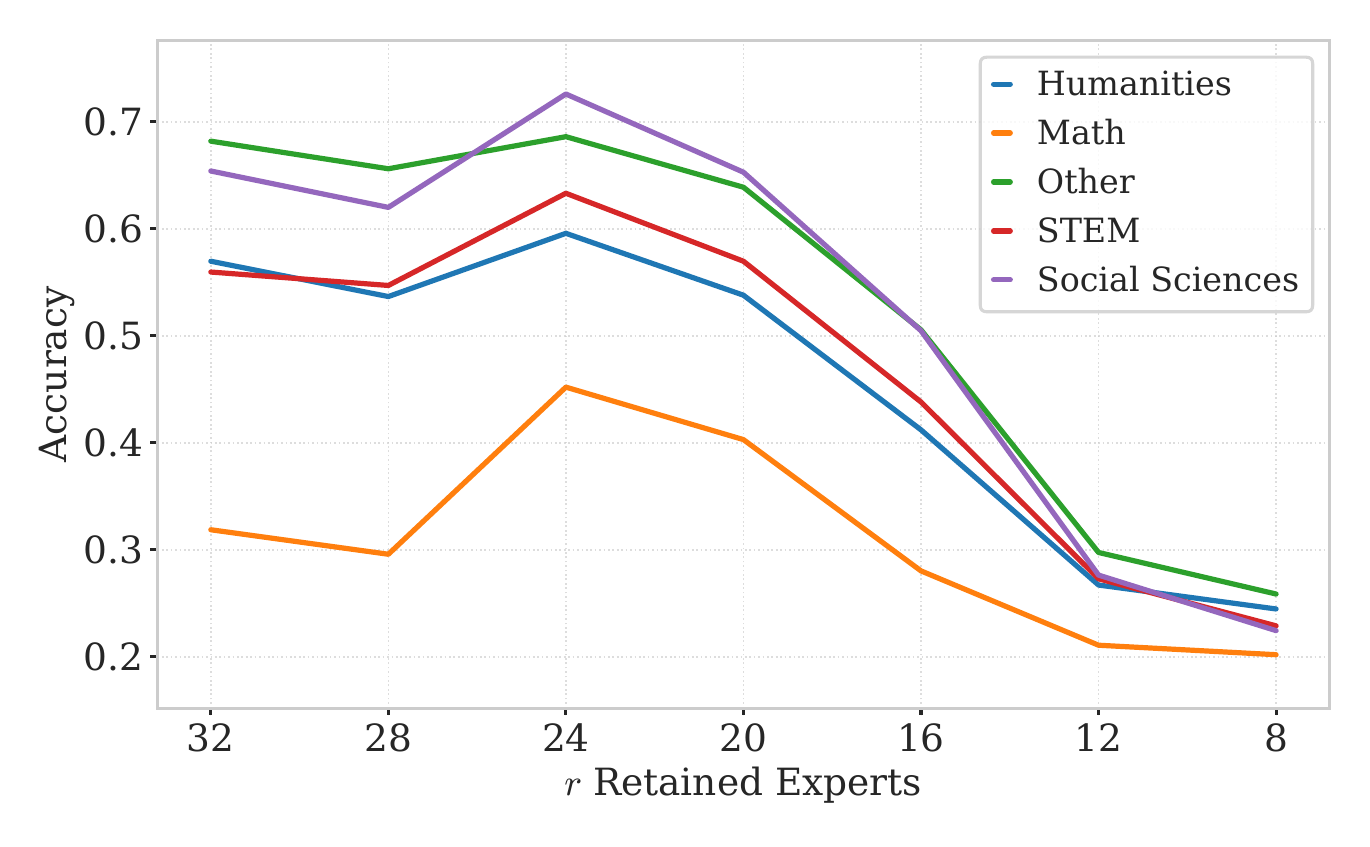}
  \caption{Evaluation performance ablation of different numbers of retained experts.}
  \label{fig:exp}
\end{figure}

\begin{figure*}[t]
  \includegraphics[width=\textwidth]{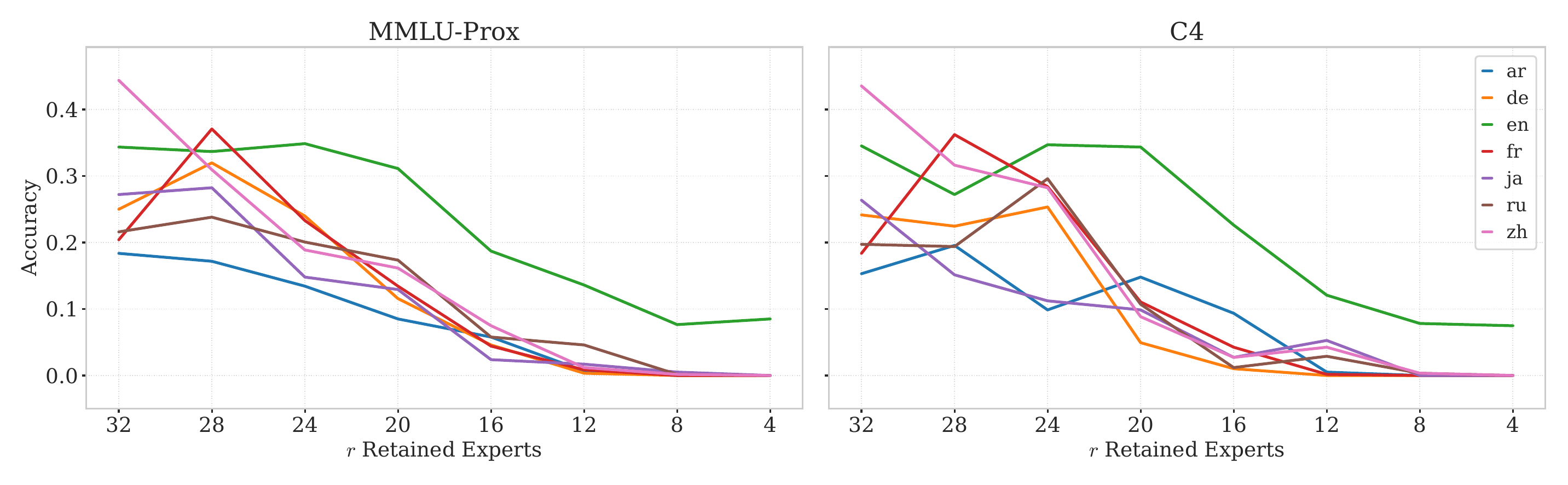}
  \caption{Evaluation performance across different languages with the language-specific MMLU-ProX calibration (left) and English C4 calibration (right) datasets. Each line represents one of 7 typologically diverse languages calibrated with English-only data.}
  \label{fig:multi}
\end{figure*}

\subsection{Optimal Set Size}

Figure~\ref{fig:exp} reports downstream performance as a function of the fraction of experts retained.
Across tasks, we observe that an optimal region that retains roughly three-quarters of the experts yields optimal performance, in some cases slightly exceeding the unpruned baseline.
To match the baseline performance, we can remove approximately half of the experts. Beyond this point, performance degrades more sharply as additional experts are removed.

Notably, we observe a consistent pruning profile across diverse subject categories, with all domains exhibiting a similar optimal retention point. This uniformity suggests that expert redundancy may be a structural property of the model architecture rather than being highly subject-specific. However, this observation may be influenced by our use of domain-general calibration data, which inherently prioritizes general experts over specialists. It is possible that strictly domain-specific calibration or a broader taxonomy of evaluation tasks could reveal more distinct expert requirements.

\subsection{Calibration Set Size}

Intuitively, a small amount of data wouldn't capture much of the distribution we care about, leading to suboptimal pruning. However, figure~\ref{fig:calib} shows that increasing the calibration set size does not consistently improve post-pruning performance once a small number of calibration sequences is available. This is useful in deployment settings because calibration data can be costly to collect, privacy-sensitive, or unavailable in the target language. Our results suggest that, for domain-general pruning in the moderate compression regime, calibration primarily needs to provide a stable estimate of broadly useful expert pathways rather than exhaustively cover the downstream distribution. 

Standard MoE training relies heavily on auxiliary load-balancing losses to ensure high hardware utilization. While necessary for training stability, these losses actively discourage true specialization by forcing the router to dispatch tokens to sub-optimal experts merely to satisfy capacity constraints. In practice, this results in redundant experts and significant functional overlap. The flat trend we observe could indicate that the experts are orthogonal post-pruning, which possibly means that pruning experts could improve the model's orthogonality. In that case, a domain-general calibration set that captures the pretraining distribution may be an optimal choice to eliminate the redundancy without overfitting towards a specific downstream task. 

\subsection{Cross-lingual Calibration}

Figure~\ref{fig:multi} evaluates whether the calibration language must match the evaluation language. In the moderate retained-expert region, English calibration remains competitive across the seven evaluated languages. Under more aggressive pruning, however, cross-lingual calibration degrades faster, suggesting that language mismatch may becomes more important when fewer experts remain.

This result is useful for multilingual deployment. In many production settings, high-quality calibration data may be readily available in English but limited or privacy-sensitive in the target language. Our findings suggest that English calibration can be a practical first-pass choice for domain-general multilingual pruning, provided that the final retained-expert ratio is validated on target-language tasks. The result also suggests that the expert-importance signal for the reasoning- and knowledge-oriented tasks we evaluate is not dominated by surface language alone. We can also infer that the model's orthogonality is largely based on concepts rather than languages which aligns with previous work on multilingual alignment on both dense and MoE models \cite{Salim_Ku_Pao_2026, Bandarkar2025-tv}. This behavior is valuable for multilingual modelling and pretraining, ensuring efficient cross-lingual transfer of knowledge and tasks. Direct analysis of language-specific routing and expert specialization remains future work.

\subsection{Calibration Distribution}

\begin{table}[t]
\centering
\small
\begin{tabular}{l l c c c}
\toprule
\textbf{Model} & \textbf{Calibration} & \textbf{MathQA} & \textbf{MMLU} & \textbf{Avg.} \\
\midrule
baseline & N/A & 0.365 & 0.563 & 0.464 \\
\midrule
\multirow{4}{*}{pruned}
& MATH    & \textbf{0.408} & 0.585 & 0.497 \\
& C4      & 0.3889 & \textbf{0.617} & 0.503 \\
& Private & 0.405  & 0.611 & \textbf{0.508}\\

\bottomrule
\end{tabular}
\caption{MathQA and MMLU performance after pruning with different calibration sets.}
\label{tab:mathqa_mmlu_results}
\end{table}

Table~\ref{tab:mathqa_mmlu_results} shows our preliminary pruning outcomes when calibrating on different data distributions.
We find that a pretraining-like corpus (C4) can be competitive with, and in some cases outperform, task-specific calibration (MATH) on downstream benchmarks. We observe that our private data performs comparably across the two categories, even though it is domain-general, yielding results similar to those from the math-specific calibration. This result aligns with findings from \citet{ji2025bewarecalibrationdatapruning}, which indicate that domain-general calibration data, similar to the pretraining distribution, are superior.

We can also observe how different data distributions affect the degradation rate in a cross-lingual setup, as shown in Figure~\ref{fig:multi}. Although subtle, the result of domain-general calibration (C4) yields slower degradation across all languages than that of task-specific calibration, as measured on the English MMLU ProX validation set. Interestingly, domain-general calibration also improves performance in the optimal range, approaching or matching English-language performance. 

\subsection{Calibration Distribution}

Table~\ref{tab:mathqa_mmlu_results} compares pruning outcomes under task-specific, public pretraining-like, and proprietary diversity-oriented calibration. The task-specific MATH calibration obtains the highest MathQA score, while C4 obtains the highest MMLU score. The proprietary corpus is not the best on either individual benchmark, but it achieves the strongest unweighted average across MathQA and MMLU.

A narrow calibration can favor the evaluation distribution it resembles, whereas a broad calibration may better preserve capabilities across unrelated tasks. The proprietary corpus was constructed to have broad domain coverage, quality filtering, and discourse variation. Its balanced performance suggests that diversity can serve as a practical proxy for estimating broadly useful experts when the deployment target spans multiple domains or domain-general. While capturing the pretraining data distribution completely may not always be possible in practice due to the massive use of proprietary data even in open-weights models, we demonstrate how data engineering strategies can be employed to construct a good proxy of this distribution

\section{Conclusion}

We presented an empirical study of the operational robustness of post-training expert pruning for MoE language models.
Across ablations over retained-expert ratios, calibration set sizes, calibration languages, and calibration distributions, we find that pruning decisions are substantially more robust than commonly assumed in deployment practice. In particular, pruning quality doesn't necessarily increase as we increase calibration set size, cross-lingual calibration can remain competitive in the regime that matters for deployment, and pretraining-like public calibration data can be a strong proxy for task-specific or private data. While fully capturing the pretraining distribution is not always possible, we define two data engineering practices to expand the diversity of raw web data.

These findings suggest a practical recipe for domain-general expert pruning. We show how domain-general pruning can help reduce redundancy and ensure an efficient deployment. Beyond immediate operational benefits, these findings suggest that the orthogonality of experts aligns across languages. Opening interesting questions about the distribution of languages and expert orthogonality in MoE models, all of which will provide valuable insights for a more efficient standard or multilingual MoE models.

\section{Future Works}

Future work should focus on why the observed phenomena happened and how they are connected with the MoE architecture specifically. The nature of basic pruning and how it interacts with sparsely activated parameters is largely under-discussed and is a promising direction given the rising popularity of MoE models in production.

\section{Limitations}

Our results are limited to a set of reasoning- and knowledge-oriented benchmarks and a fixed pruning policy. Furthermore, the study is limited to a single MoE model architecture and size due to compute limitations. Verifying transferability to models with different architectures and scales remains a promising direction.

\bibliography{custom,anthology-1,anthology-2}
\clearpage
\appendix

\section*{Appendix}

\section{Different Domains Mapping}
\label{sec:appendix}

\begin{table}[htpb]
\centering
\small
\setlength{\tabcolsep}{4pt} 
\renewcommand{\arraystretch}{1.2} 

\begin{tabularx}{\columnwidth}{@{} l >{\raggedright\arraybackslash}X @{}}
\toprule
\textbf{Category} & \textbf{Labels} \\
\midrule

\textbf{Math}
& gsm8k, abstract\_algebra, college\_mathematics, elementary\_mathematics, high\_school\_mathematics, high\_school\_statistics \\
\midrule

\textbf{STEM}
& anatomy, astronomy, college\_biology, college\_chemistry, college\_computer\_science, college\_physics, computer\_security, conceptual\_physics, electrical\_engineering, high\_school\_biology, high\_school\_chemistry, high\_school\_computer\_science, high\_school\_physics, machine\_learning \\
\midrule

\textbf{Humanities}
& humanities, formal\_logic, high\_school\_european\_history, high\_school\_us\_history, high\_school\_world\_history, international\_law, jurisprudence, logical\_fallacies, moral\_disputes, moral\_scenarios, philosophy, prehistory, professional\_law, world\_religions \\
\midrule

\textbf{Social Sciences}
& social\_sciences, econometrics, high\_school\_geography, high\_school\_government\_and\_politics, high\_school\_macroeconomics, high\_school\_microeconomics, high\_school\_psychology, human\_sexuality, professional\_psychology, public\_relations, security\_studies, sociology, us\_foreign\_policy \\
\midrule

\textbf{Other}
& other, business\_ethics, clinical\_knowledge, college\_medicine, global\_facts, human\_aging, management, marketing, medical\_genetics, miscellaneous, nutrition, professional\_accounting, professional\_medicine, virology \\
\bottomrule
\end{tabularx}

\caption{Grouping of MMLU subtasks used in visualizations for brevity, with GSM8K as an addition in the Math category.}
\label{tab:mmlu_grouping}
\end{table}

For brevity, the results shown in this study group multiple domains of several subtasks with the mappings provided in Table~\ref{tab:mmlu_grouping}.

\end{document}